\documentclass{article}
\usepackage{arxiv}

\usepackage{amsmath,amssymb}
\usepackage{algorithm}
\usepackage{algpseudocode}
\usepackage{graphicx}
\usepackage{xcolor}
\usepackage{multirow}
\usepackage{hyperref}
\usepackage{soul}
\usepackage{booktabs}
\usepackage{pifont}
\usepackage[numbers,sort&compress]{natbib}
\usepackage{url}

\hypersetup{
    colorlinks=true,
    linkcolor=blue,
    citecolor=blue,
    filecolor=blue,
    urlcolor=blue
}

\newcommand{\cmark}{\ding{51}} 
\newcommand{\xmark}{\ding{55}} 

\title{BlendedNet: A Blended Wing Body Aircraft Dataset and Surrogate Model for Aerodynamic Predictions}

\author{
  Nicholas Sung\thanks{Equal contribution.} \\
  Department of Mechanical Engineering\\
  Massachusetts Institute of Technology\\
  Cambridge, MA\\
  \texttt{nicksung@mit.edu}
  \And
  Steven Spreizer\footnotemark[1]\\
  MIT Lincoln Laboratory\\
  Lexington, MA\\
  \texttt{steven.spreizer@ll.mit.edu}
  \And
  Mohamed Elrefaie\\
  Department of Mechanical Engineering\\
  Massachusetts Institute of Technology\\
  Cambridge, MA\\
  \texttt{mohamed.elrefaie@mit.edu}
  \And
  Kaira Samuel\\
  Center for Computational Science \& Engineering\\
  Massachusetts Institute of Technology\\
  Cambridge, MA\\
  \texttt{kmsamuel@mit.edu}
  \And
  Matthew C.\ Jones\\
  MIT Lincoln Laboratory\\
  Lexington, MA\\
  \texttt{matthew.jones@ll.mit.edu}
  \And
  Faez Ahmed\\
  Department of Mechanical Engineering\\
  Massachusetts Institute of Technology\\
  Cambridge, MA\\
  \texttt{faez@mit.edu}
}

\begin{document}
\maketitle

\begin{abstract}
We introduce \textbf{BlendedNet}, a publicly available aerodynamic dataset featuring 999 unique blended wing body (BWB) geometries. Each geometry was simulated across approximately 9 distinct aerodynamic cases, resulting in a total of 8,830 successfully converged cases. BlendedNet geometries are systematically generated using sampling across geometric design parameters and flight conditions, and analyzed with high-fidelity Reynolds-Averaged Navier-Stokes (RANS) simulations employing the Spalart-Allmaras turbulence model using 9 to 14 million volume cells per case. The dataset is publicly available at \href{https://doi.org/10.7910/DVN/VJT9EP}{Harvard Dataverse}. In addition to this dataset, we introduce a fully end-to-end surrogate modeling framework for pointwise aerodynamic coefficient prediction \( (C_p, C_{f_x}, C_{f_z}) \) which contributes to lift and drag. This framework consists of two separate models: (1) a permutation-invariant PointNet model which predicts geometric design parameters from sampled point clouds, and (2) a Feature-wise Linear Modulation (FiLM) network that takes these predicted parameters, along with flight conditions, to generate pointwise aerodynamic coefficient predictions. Our evaluations demonstrate that the surrogate model achieves accurate pointwise aerodynamic performance predictions with low errors. BlendedNet addresses critical data scarcity in the field, enabling future research into data-driven surrogate modeling methods for complex BWB aircraft aerodynamic design. 

\end{abstract}


\section{Introduction}

Blended Wing Body (BWB) aircraft have gained significant attention as a promising configuration for next-generation transport due to their potential for enhanced aerodynamic efficiency and fuel economy. The integration of the fuselage and wings into a single lifting surface allows for higher lift-to-drag ratios, reduced structural weight, and lower fuel consumption compared to conventional tube-and-wing (TAW) configurations~\cite{BoeingBWB}. Early studies on BWB aerodynamic analysis using wind tunnel testing indicated that BWB aircraft could reduce fuel costs by up to 30\% compared to traditional large aircraft, while also benefiting from reduced structural weight \cite{BoeingBWB, carter2009blended}, and current research verifies these improvements \cite{zhenli2019assessment}. An important aspect of realizing these BWB designs is the ability to accurately and efficiently model their complex aerodynamics. In recent years, the rise of computational methods in aircraft design has led to more efficient and robust aerodynamic analysis. At the same time, it has presented new challenges in accurately representing complex geometries, such as the BWB, and optimizing these designs using Computational Fluid Dynamics (CFD). 

BWB geometries exhibit more complex flow interactions than conventional TAW aircraft, making both experimental and computational analysis more demanding. Even in the early design stages, effective BWB analysis requires sophisticated parameterization methods to capture the smooth blending between body and wing, along with high-fidelity simulations to resolve the complex flow physics that occurs over the integrated lifting body \cite{HighFidelityCFD, MultiFidelityBWB}. This leads to significant slow-downs in the design process, highlighting the need for faster and more efficient design optimization methods. While researchers have developed geometric parameterizations \cite{zhang2016conceptual, GeometricModelBWB} and introduced computational methods such as multi-fidelity modeling \cite{feldstein2020multifidelity, MultiFidelityBWB, sarkar2019multifidelity, charisi2025multi} and multidisciplinary optimization \cite{HighFidelityCFD, martins2013multidisciplinary, lyu2014aerodynamic} to accelerate the process, these techniques still rely heavily on computationally expensive physics-based simulations. As a result, design iterations remain time-intensive, limiting the speed at which new configurations can be explored. Additionally, existing work primarily focuses on integrated aerodynamic coefficients (total lift and drag) or provides only sparse point-wise data at select locations \cite{loeser2010saccon, cobb2023aircraftverse, edwards2021design}, making it difficult to capture the full aerodynamic complexity of BWB aircraft. The scarcity of publicly available, high-fidelity datasets that provide detailed surface-level pressure and force distributions further exacerbates this challenge, as such data is essential for accurate modeling and optimization of these unconventional designs \cite{GeometricModelBWB}.

Data-driven approaches, which remain underexplored for BWB aircraft, offer a promising solution. Methods such as machine learning-enabled multi-fidelity frameworks have been proposed to reduce computational cost and improve surrogate accuracy in complex system design optimization tasks \cite{garriga2019machine}. In the aerospace domain, deep learning methods have shown promise for traditional aircraft designs, with researchers developing models for aerodynamic prediction using Gaussian processes, artificial neural networks (ANNs), and graph-based neural networks (GNNs) \cite{secco2017artificial, dong2021deep}. However, these approaches heavily rely on large, high-fidelity datasets and are often limited by sparse data availability and a lack of explicit surface-level information. Consequently, the absence of comprehensive, publicly available datasets for BWB configurations has restricted further exploration of data-driven design optimization methods for these aircraft.

To address data scarcity and design optimization limitations in BWB research, we introduce a publicly available high-resolution dataset specifically designed for BWB aerodynamic analysis, featuring detailed pointwise surface aerodynamic coefficients, based on the parameterization defined in \cite{zhang2016conceptual}. Additionally, we develop a deep learning framework that predicts aerodynamic surface properties in two stages. Firstly, we used a PointNet based regressor to predict geometric shape parameters based on the BWB's pointcloud. Secondly, we combined the predicted parameters with flight conditions to condition a feature-wise linear modulation (FiLM) network for pointwise predictions of aerodynamic coefficients \(C_p, C_{f_x}, C_{f_z}\) which contributes to lift and drag across the aircraft surface. Thus, the key contributions of this paper include:
\begin{enumerate}
    \item \textbf{A High-Resolution Surface Aerodynamic Dataset} – The first publicly available dataset with detailed pointwise aerodynamic coefficients for BWB configurations, generated using high-fidelity Reynolds-Averaged Navier-Stokes (RANS) simulations with the Spalart-Allmaras turbulence model. The dataset encompasses both shape/geometry variations and flight condition variations, providing a comprehensive benchmark for aerodynamic analysis and surrogate modeling. We release BlendedNet as an open dataset at \href{https://doi.org/10.7910/DVN/VJT9EP}{Harvard Dataverse}.

    \item \textbf{A Fully End-to-End Surrogate Model for Pointwise 3D Surface Predictions } – A deep learning framework that predicts aerodynamic surface properties in two stages. (1) A permutation-invariant PointNet model first predicts geometric design parameters from a sampled point cloud. (2) These predicted parameters, along with flight conditions, then condition a Feature-wise Linear Modulation (FiLM) network, which predicts pointwise aerodynamic coefficients for the aircraft surface. This pipeline enables seamless aerodynamic surface predictions directly from a sampled point cloud and flight conditions.
\end{enumerate}

\section{Related Work}
The Blended Wing Body (BWB) configuration has been studied for its potential to improve aircraft design by significantly reducing fuel consumption and mitigating environmental concerns, such as emissions and ground noise.Before advances in CFD analysis, the advantages of the BWB configuration were validated through extensive wind tunnel testing, demonstrating increased lift-to-drag ratios and improved structural efficiency due to lighter, yet more durable designs \cite{BoeingBWB, carter2009blended}. Now, the transition to incorporating high-fidelity computational methods has improved the efficiency and robustness of aerodynamic validation \cite{catalani2024neural}. However, even in the conceptual design phase, high-fidelity aerodynamic analysis is necessary to accurately capture the intricate aerodynamics of BWB configurations due to the unconventional geometry\cite{qin2004aerodynamic}. This provides an opportunity to explore data-driven methods for design optimization using high-fidelity  BWB data. The following sections discuss prior work on the conceptual design of BWB as well as machine learning methods for aerodynamic prediction in aerospace applications.

\paragraph{Design Optimization Methods}

Researchers have explored various methods to optimize BWB designs efficiently. Geometric parameterization techniques have been used to explore early-stage design. For example, Mastiny et al. developed a parametric model of the BWB aircraft from a preliminary CAD geometry, specifically designed to facilitate conceptual design optimization, allowing systematic modifications of BWB shapes while maintaining aerodynamic performance \cite{GeometricModelBWB}. 

Beyond geometric parametrization, different computational design optimization methods have been proposed to balance computational cost and accuracy, including multi-fidelity simulations and multidisciplinary design optimization (MDO). Multi-fidelity optimization approaches leverage low-fidelity models for rapid iteration while incorporating high-fidelity simulations for final validation \cite{MultiFidelityBWB, charisi2025multi, sarkar2019multifidelity, feldstein2020multifidelity}. Additionally, MDO has been explored for integrating aerodynamic, structural, and propulsion considerations into a cohesive and more efficient design process \cite{zhang2016conceptual, HighFidelityCFD, martins2013multidisciplinary, lyu2014aerodynamic}. Unlike sequential design methodologies historically used in commercial aircraft development, MDO facilitates simultaneous optimization across multiple disciplines, which is particularly critical for unconventional configurations such as the BWB. Existing research utilizing MDO has explored high-fidelity simultaneous optimization across trim, stability, and structural constraints for later-stage design  \cite{lyu2014aerodynamic, HighFidelityCFD}, as well as lower-order physics modeling that employs multi-objective genetic algorithms rather than traditional gradient-based methods for earlier-stage design \cite{zhang2016conceptual}. 

Despite these considerable advances in BWB design optimization, most approaches remain computationally intensive and time-consuming, particularly when high-fidelity analysis is required. The exploration of data-driven methods in conventional aircraft design suggests that these methods could continue improving the efficiency of BWB design by dramatically reducing the computational burden of early-stage exploration and optimization.

\paragraph{Machine Learning for Aerodynamic Prediction}
The lack of comprehensive, publicly available datasets for BWB aircraft has left machine learning (ML) approaches largely unexplored in BWB design optimization, despite their proven success in other aerospace applications. One prior work developed a neural network for the inverse design of the BWB aircraft, where predictions of design variables, including wingspan, wing area, maximum takeoff weight, and thrust available, were derived from aerodynamic performance quantities, including lift and drag \cite{sharma2024mission}. While this demonstrates the potential for ML in the inverse problem, there remains a critical need to solve the more challenging forward problem of accurately predicting detailed aerodynamic properties from geometry. 

ML methods have been proposed as a promising approach to general aerodynamic prediction in other aircraft designs, addressing some of the limitations of traditional physics-based methods. Traditional methods often struggle with handling uncertainty and noise in response to changing operational conditions \cite{dong2021deep}. Furthermore, they are highly computationally expensive and time-intensive, indicating a gap that can be optimized in the design process \cite{sabater2022fast}. These challenges can be tackled through a bottom-up, data-driven approach to design \cite{dong2021deep, sabater2022fast, garriga2019machine}. In particular, machine learning has shown great potential in the aircraft domain, with Gaussian processes \cite{sabater2022fast, chen2021framework, liu2014modeling} and artificial neural networks \cite{secco2017artificial} being used for aircraft aerodynamic prediction.  Deep learning models, including convolutional neural networks (CNNs) and deep autoencoders (DAEs) have also been explored, especially for more complex, three-dimensional data. Graph neural networks (GNNs) have also been studied for their compatibility with mesh-like data, which is a common data representation in the aerospace domain \cite{dong2021deep}. 

Neural fields have also been explored, which propose a coordinate-based neural network formulation that operates on unstructured domains, allowing for increased generalization to unseen data \cite{catalani2024neural}. Their work reported test error that was more than three times lower than the test error in GNNs, suggesting an impressive improvement over traditional deep learning methods. With point-wise aerodynamic coefficient data, this method shows a lot of promise for the BWB configuration. 

\paragraph{Limitations of Current Work}
While current research has begun addressing the computational limitations that come with aerodynamic analysis of BWB configurations, there is an opportunity for major speed-ups using data-driven methods. However, many of these methods heavily rely on large, high quality datasets as they derive their predictive capabilities entirely from the data. Recent advancements have demonstrated the potential of scaling dataset size to enhance surrogate model development and improve deep learning predictions in 3D engineering problems, as shown in\cite{elrefaie2025drivaernet,elrefaie2024drivaernet++}. Despite this progress, the lack of robust high-fidelity datasets for BWB aircraft, as seen in Table \ref{tab:aero_datasets}, limits the development and validation of data-driven models for this complex design space.   We believe the release of the first public high-fidelity BWB dataset, along with a machine learning method for aerodynamic predictions on the high-fidelity data, will begin addressing these gaps. The proposed method leverages Feature-wise Linear Modulation (FiLM), which employs a dynamic internal architecture that can adapt to the varying conditions and complex behavior of Blended Wing Body (BWB) designs \cite{perez2018film}. The goal of this work is to advance the state-of-the-art in BWB aerodynamic analysis by providing a comprehensive dataset and an effective predictive framework, facilitating future research in surrogate modeling.

\begin{table*}[]
    \centering
    \caption{Comparison of 3D Aircraft Aerodynamic Datasets and Case Studies}
    \label{tab:aero_datasets}
    \resizebox{\textwidth}{!}{ 
    \begin{tabular}{@{\hskip 6pt}l@{\hskip 6pt} l@{\hskip 6pt} c@{\hskip 6pt} c@{\hskip 6pt} c@{\hskip 6pt} c@{\hskip 6pt} c@{\hskip 6pt} c@{\hskip 6pt} c@{\hskip 6pt}} 
        \toprule
        \textbf{Type} & \textbf{Name} & \textbf{Size} & \multicolumn{3}{c}{\textbf{Aerodynamic Values}} & \textbf{Modalities} & \textbf{BWB} & \textbf{Open Source} \\
        \cmidrule(lr){4-6}
        & & & $C_L$ / $C_D$ & $C_m$ & Pointwise $C_p$ / $C_f$ &  &  &  \\ 
        \midrule
        \multirow{4}{*}{Case Study} & NASA BWB-450-1L \cite{carter2009blended} & 1 geometry, 27 conditions & \cmark & \cmark & \xmark & Physical & \cmark & \xmark \\
        & NASA/Boeing X-48B BWB \cite{BoeingBWB} & 2 geometries, 10+ conditions & \cmark & \cmark & \xmark & Physical & \cmark & \xmark \\
        & SACCON Dataset \cite{loeser2010saccon} & 1 geometry, 40K+ conditions & \cmark & \cmark & \xmark & Physical & \cmark & \xmark \\
        & NASA CRM\cite{rivers2019nasa} (DPW \cite{sclafani2013dpw} + ETW\cite{lutz2013going} CFD) & 1 geometry & \cmark & \cmark & \cmark & Mesh, Parametric & \xmark & Partial\textsuperscript{*} \\
        \midrule
        \multirow{3}{*}{Dataset} & AircraftVerse\cite{cobb2023aircraftverse} & 27,714 geometries & \cmark & \xmark & \xmark & Mesh, Parametric & \xmark & \cmark \\
        & ShapeNet Aircraft\cite{edwards2021design} & 4,045 geometries & \cmark & \xmark & \xmark & Mesh, 2D images & \xmark & \cmark \\
        & BlendedNet & 999 geometries, $\sim 10 \,$ conditions  & \cmark & \cmark & \cmark & Mesh, Parametric & \cmark & \cmark \\
        \bottomrule
    \end{tabular}
    }
    \vspace{0.5em}
    \footnotesize{\textsuperscript{*}Select mesh files (`.bdf`) and structural results (`.f06`) are available, but full aerodynamic datasets (e.g., full CFD or wind tunnel data) are not openly accessible.}
\end{table*}

\section{Dataset Generation}
The primary objective of this work is to develop a publicly accessible dataset of aerodynamic simulations for blended wing body (BWB) aircraft, enabling the exploration of machine learning models for aerodynamic performance prediction. To the best of the authors’ knowledge, no such dataset currently exists.

The BWB configuration was chosen due to growing interest in its potential for improved aerodynamic efficiency and fuel savings \cite{BoeingBWB, zhenli2019assessment}, as well as the diverse range of geometries that can be generated through its parameterization. The dataset includes both integrated force coefficients (1D data) and distributed surface coefficients (2D data). However, full flowfield solutions (3D data) were excluded to manage data volume constraints while maintaining practical usability.

\paragraph{Geometry Generation}

\begin{figure}
    \centering
    \includegraphics[width=0.7\linewidth]{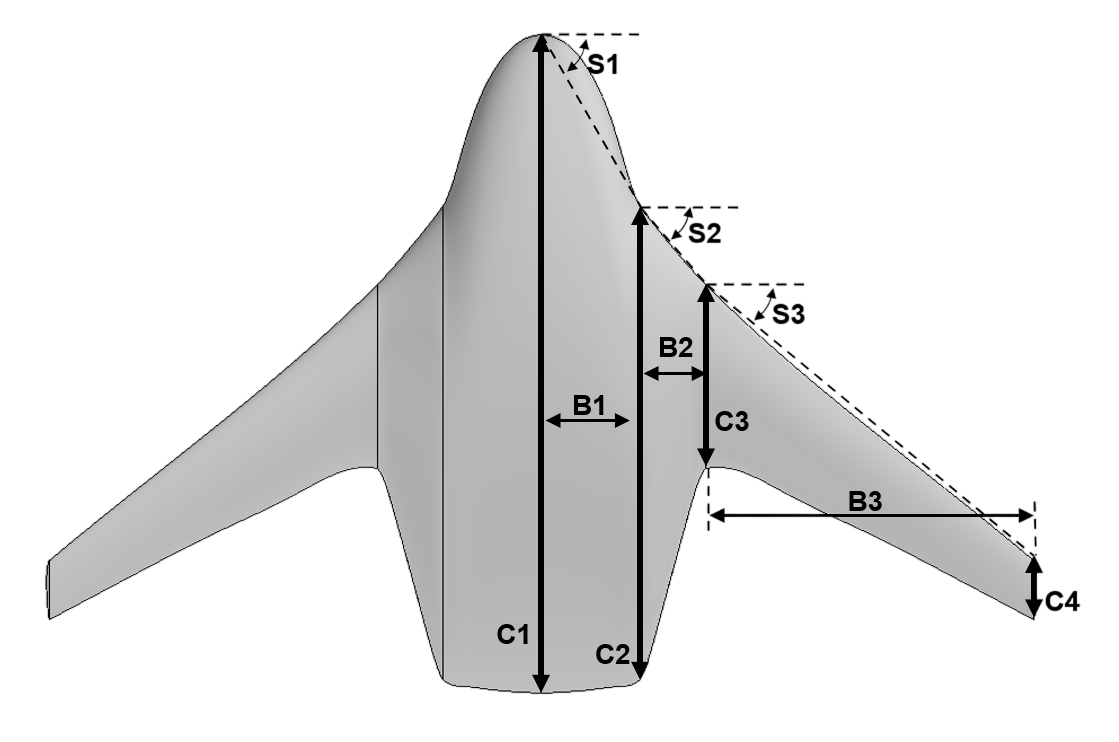}
    \caption{Illustration of the blended wing body (BWB) planform parameterization, showing key geometric design parameters used in dataset generation.}
    \label{fig:parameter_location}
\end{figure}

A parameterized blended wing body (BWB) aircraft was modeled using the Open Vehicle Sketch Pad (OpenVSP) \cite{OpenVSP} software. The selected parameterization primarily focuses on the aircraft planform and is inspired by the work of Zhang et al.\cite{zhang2016conceptual}. The definitions of the various geometric design parameters are illustrated in \autoref{fig:parameter_location}, while \autoref{tab:geometry_parameters} provides an overview of the parameter ranges, normalized by the centerline length, which is selected as the Reynolds length.

To generate the dataset, Latin Hypercube Sampling (LHS) was employed to systematically sample the planform parameters, resulting in a total of 999 unique geometries.
The airfoil cross-sections were parameterized using a degree-4 Class-Shape Transformation (CST) for the upper and lower surfaces.
For this study, airfoil parameters were held constant to reduce the dimensionality of the problem; however, future work may explore variations in airfoil shapes. Histograms of the geometric parameter distributions are provided in the Appendix (see \autoref{fig:geom_param_dist}).

Once the OpenVSP models were generated, they were imported into Engineering Sketch Pad (ESP) \cite{ESP} to facilitate CAD export in STEP (.stp) format. 
ESP was chosen for its robust solid modeling capabilities, including Boolean operations, which are not natively supported in OpenVSP. 
A CAD union operation was performed to merge the sections of the aircraft model before exporting the final geometry for the subsequent meshing step.

\begin{table}[h]
    \centering
    \caption{Design parameter ranges for the blended wing body (BWB) aircraft geometry.}
    \label{tab:geometry_parameters}
    \begin{tabular}{@{} ccc c @{}} 
        \hline
        \textbf{Design Parameter} & \textbf{Units} & \textbf{Lower Bound} & \textbf{Upper Bound} \\
        \hline
        C2/C1 & - & 0.55 & 0.85 \\
        C3/C1 & - & 0.18 & 0.28 \\
        C4/C1 & - & 0.06 & 0.09 \\
        B1/C1 & - & 0.10 & 0.20 \\
        B2/C1 & - & 0.05 & 0.20 \\
        B3/C1 & - & 0.20 & 0.70 \\
        S1 & deg & 40 & 60 \\
        S2 & deg & 40 & 60 \\
        S3 & deg & 24 & 40 \\
        \hline
    \end{tabular}
\end{table}

\paragraph{Meshing}

\begin{figure}[h]
    \centering
    \includegraphics[width=0.7\linewidth]{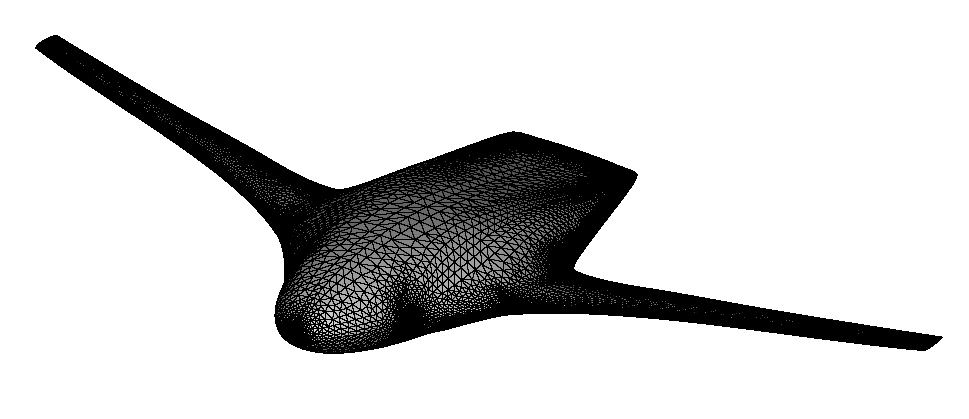}
    \caption{View of an example BWB surface mesh.}
    \label{fig:mesh_iso}
\end{figure}

\begin{figure}[h]
    \centering
    \includegraphics[width=0.7\linewidth]{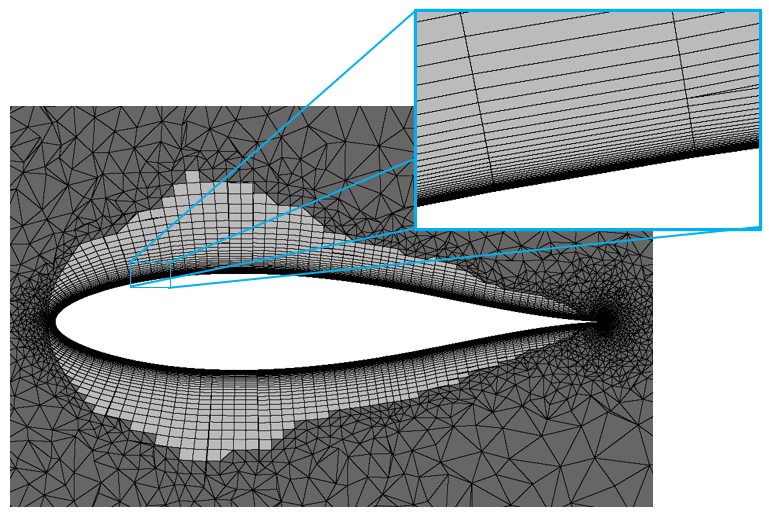}
    \caption{Centerline slice of mesh with zoomed view of boundary layer cells.}
    \label{fig:mesh_slice}
\end{figure}

Computational meshes were generated using the Pointwise CFD meshing software ~\cite{pointwise}.
To automate the meshing process, scripts were developed using the Pointwise Glyph scripting language, which is based on TCL 2.
The generated surface meshes consisted of 34,729 to 85,392 points, depending on the specific geometry.
\autoref{fig:mesh_iso} shows an example BWB surface mesh.
The corresponding volume meshes contained between 9 and 14 million cells and comprised a mix of element types, including tetrahedra, hexahedra, prisms, and pyramids.
Boundary layer cells were grown off the surfaces to fully resolve the near-wall flow, with wall-normal spacing selected to maintain \( y^+ \)  values below 1. \autoref{fig:mesh_slice} illustrates a centerline slice of a BWB CFD mesh and highlights the resolution of the boundary layer cells.

\paragraph{CFD Simulation}

Computational fluid dynamics (CFD) simulations were conducted using NASA's FUN3D solver~\cite{anderson2024fun3d}. The simulations were performed to reach a steady-state solution by solving the Reynolds-Averaged Navier-Stokes (RANS) equations, employing the Spalart-Allmaras turbulence model for closure.  The Spalart-Allmaras turbulence model was employed due to its numerical robustness, well-behaved performance in boundary layers, minimal sensitivity to freestream values, and ease of implementation in compressible flow solvers. ~\cite{allmaras2012modifications}.

Flight conditions shown in \autoref{tab:flightconditions} were sampled using Latin Hypercube Sampling (LHS) over key parameters: altitude, Mach number, angle of attack, and Reynolds length. The Reynolds number was not directly sampled; instead, it was computed during post-processing using the freestream properties and the specified Reynolds length. Reference area and lengths for force and moment non-dimensionalization were consistently set to 1 across all geometries. A far-field spherical boundary was used to define inflow and outflow conditions in the simulation domain, while a no-slip boundary condition was applied at the wall surfaces. Histograms of the sampled flight condition distributions are provided in the Appendix (see \autoref{fig:flight_conditions}).

\begin{table}[]
    \centering
    \caption{Flight condition ranges used to generate CFD cases.}
    \label{tab:flightconditions}
    \begin{tabular}{@{}ccc@{}}
        \hline
         \textbf{Parameter} & \textbf{Lower Bound} & \textbf{Upper Bound}  \\ \hline
         Altitude [kft] & 0&40 \\
         Mach [-] & 0.05& 0.5\\
         Reynolds Length [m] & 0.1& 10\\
         Angle of Attack [deg] & -10 & 20 \\ \hline
    \end{tabular}
\end{table}

\paragraph{Results Processing}
Post-processing of the simulation results involved filtering out CFD cases where the residuals did not sufficiently converge within a fixed number of timesteps (convergence defined as density residual below $1\times10^{-5}$ and turbulence residual below $1\times10^{-4}$).
Only successfully converged simulations were retained, with outputs including the integrated lift, drag, and pitching moment coefficients. 
For consistency, the pitching moment was referenced at the vehicle nose.  
The simulation data was then aggregated with the corresponding input geometry parameters and flight conditions to construct a comprehensive results table containing all relevant scalar quantities. 
Additionally, surface pressure coefficients (\( C_p \)) and skin friction components (\( C_{f_x} \), \( C_{f_y} \), \( C_{f_z} \)) were converted to the open-source VTK format\cite{vtkBook} for visualization and further analysis.

\begin{figure*}[h]
    \centering
    \includegraphics[width=\textwidth]{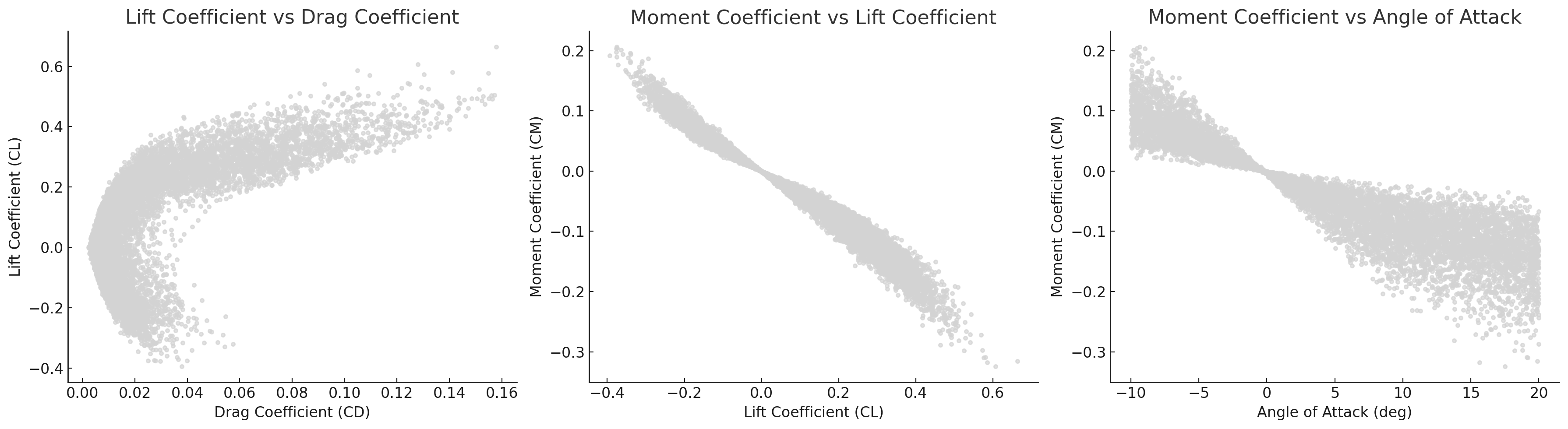}
    \caption{Scatter plots of aerodynamic coefficients: (Left) Lift coefficient ($C_L$) vs Drag coefficient ($C_D$), (Center) Pitching moment coefficient ($C_M$) vs Lift coefficient ($C_L$), and (Right) Pitching moment coefficient ($C_M$) vs angle of attack ($\alpha$). These relationships provide insights into aerodynamic performance and longitudinal stability characteristics.}
    \label{fig:cl_cd_cm}
\end{figure*}

\section{Dataset Characteristics}

\autoref{tab:generation_cost} details the computational time required to produce the dataset in this report.
Due to software availability, the mesh generation and post-processing steps were performed on a Windows workstation while the other steps were run on a Linux based cluster.
Cluster resources were provided by the MIT Lincoln Laboratory Supercomputing Center \cite{reuther2018interactive}.
The final dataset (uncompressed) is roughly 50 GB, with each VTK file being on average 5-7 MB.

\begin{table}[]
    \centering
    \caption{Summary of computational cost per step of the data generation workflow.}
    \label{tab:generation_cost}
    \begin{tabular}{@{}ccc@{}}
        \textbf{Task}&\textbf{Compute Resource} & \textbf{Time per Case} \\ \hline
         Geometry Generation&Xeon-p8 CPU&  10-15 sec\\
         Mesh Generation& i9 Workstation& 3-4 min\\
         CFD Solve & 2xV100 GPU & 1.5 hr\\
         Post-Processing &i9 Workstation& 5-10 sec \\ \hline
    \end{tabular}
\end{table}

Integrated quantities of $C_L$, $C_D$, and $C_{M_y}$ can be recovered from the point-wise distributions of this dataset.
Distributed pressure and skin friction forces can aggregated into the body frame via
\begin{equation}
    C_i = -C_p \hat{n}_i + C_{f_i},\ \ i\in [x,y,z]
\end{equation}
where $\hat{n}_i$ is the $i$ component of the local normal vector, $C_p$ is the local coefficient of pressure, and $C_{f_i}$ is the $i$ component of local skin friction coefficient.
To transform the forces into the reference frame of the flow:

\begin{equation}
    C_d=C_x \text{cos}(\alpha) + C_z\text{sin}(\alpha)
\end{equation}
\begin{equation}
    C_\ell=-C_x \text{sin}(\alpha) + C_z\text{cos}(\alpha)
\end{equation}
where $\alpha$ is the aircraft angle of attack.

Integrated force coefficients can be calculated from the point-wise values by

\begin{equation}
    C_L = \frac{1}{A_{ref}}\sum_i C_{\ell_i}A_i
\end{equation}

\begin{equation}
    C_D= \frac{1}{A_{ref}}\sum_i C_{d_i}A_i
\end{equation}
where $i$ is the index of surface panels, $A_i$ is the panel area, and $A_{ref}$ is the aircraft reference area (1 for this work).
Integrated moment coefficient can likewise be computed by
\begin{equation}
    C_{M_y} = \frac{1}{A_{ref}c_{ref}}\sum_i C_{x_i}A_i\Delta z_i - C_{z_i}A_i\Delta x_i
\end{equation}
where $\Delta x_i$ and $\Delta z_i$ are the x- and z-direction offsets from the moment reference point for the $i$th panel and $c_{ref}$ is the reference chord length for moment nondimensionalization (also 1 for this work). 

The relationships between aerodynamic coefficients, as illustrated in \autoref{fig:cl_cd_cm}, provide insights into the trade-offs between lift, drag, and pitching moment, which are critical for evaluating aerodynamic performance and stability. The plot of the lift coefficient (\( C_L \)) against the drag coefficient (\( C_D \)) typically exhibits a classic U-shaped trend. At lower values of \( C_L \), the drag \( C_D \) is relatively small. As \( C_L \) increases, the drag also increases, primarily due to the induced drag associated with generating higher lift. This behavior is expected because producing lift inherently leads to additional drag, especially at higher angles of attack where flow separation and vortex generation may occur. Secondly, the relationship between the lift coefficient (\( C_L \)) and the pitching moment coefficient (\( C_M \)) shows a strong negative correlation. As \( C_L \) increases, \( C_M \) becomes more negative, indicating a stronger balancing nose-down pitching moment. Lastly, the relationship between the pitching moment coefficient (\( C_M \)) and the angle of attack (\( \alpha \)) reveals additional insight into the longitudinal stability of the configuration. (\( C_M \)) decreases (becomes more negative) with increasing \( \alpha \). This trend suggests that the configuration exhibits \textit{static longitudinal stability}, meaning that as the nose pitches up (increasing \( \alpha \)), a restoring nose-down moment (more negative \( C_M \)) develops to bring the aircraft back toward equilibrium. The slope of the \( C_M \) vs \( \alpha \) curve is an important indicator: a negative slope typically reflects a stable design, where the aerodynamic moment resists deviations from the trim condition.

\begin{figure}[htbp]
    \centering
    \includegraphics[width=\textwidth]{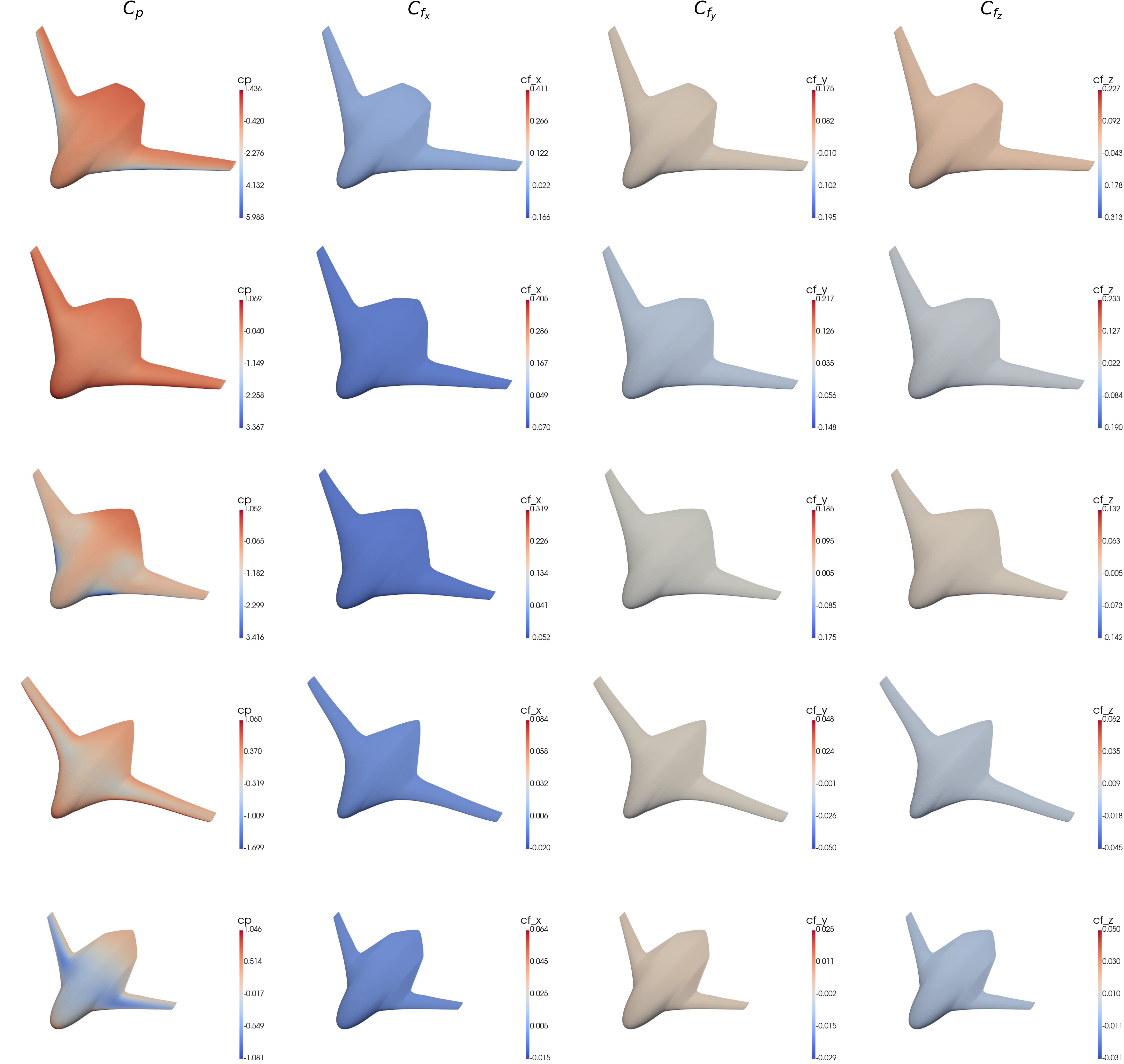}
    \caption{Visualization of aerodynamic coefficients for different cases. Each row represents a specific design and aerodynamic flight condition respectively: highest and lowest lift, highest drag, and highest and lowest lift-to-drag ratio. The columns correspond to the pressure coefficient ($C_p$) and the skin friction coefficients in the $x$, $y$, and $z$ directions ($C_{f_x}$, $C_{f_y}$, and $C_{f_z}$). It is important to emphasize that both the BWB geometry and the flight conditions must be taken into account, as their interaction jointly influences the overall aerodynamic performance.}
    \label{fig:vtk_visualization}
\end{figure}

\section{Surrogate Model Development}
Our surrogate modeling framework uses two neural networks in sequence, forming an end-to-end pipeline. 
 First, a permutation-invariant PointNet model is trained to predict geometric shape parameters from 5000 sampled points on the surface. The PointNet model is inherently permutation-invariant and excels at learning global shape features from unstructured data while maintaining permutation invariance. Next, a Feature-wise Linear Modulation (FiLM) neural network uses these predicted geometric parameters, combined with known flight conditions, to predict point wise aerodynamic surface properties. FiLM dynamically modulates network layers based on external inputs, making it effective for capturing complex, shape-dependent aerodynamic behavior across different flight conditions. ~\cite{perez2018film}. This two-stage pipeline enables prediction of the full aerodynamic field across the surface, using only sampled surface points and flight conditions as input.

\subsection*{PointNet Model for Predicting Geometric Design Parameters from Point Clouds}

\begin{figure}[htbp]
    \centering
    \includegraphics[width=\textwidth]{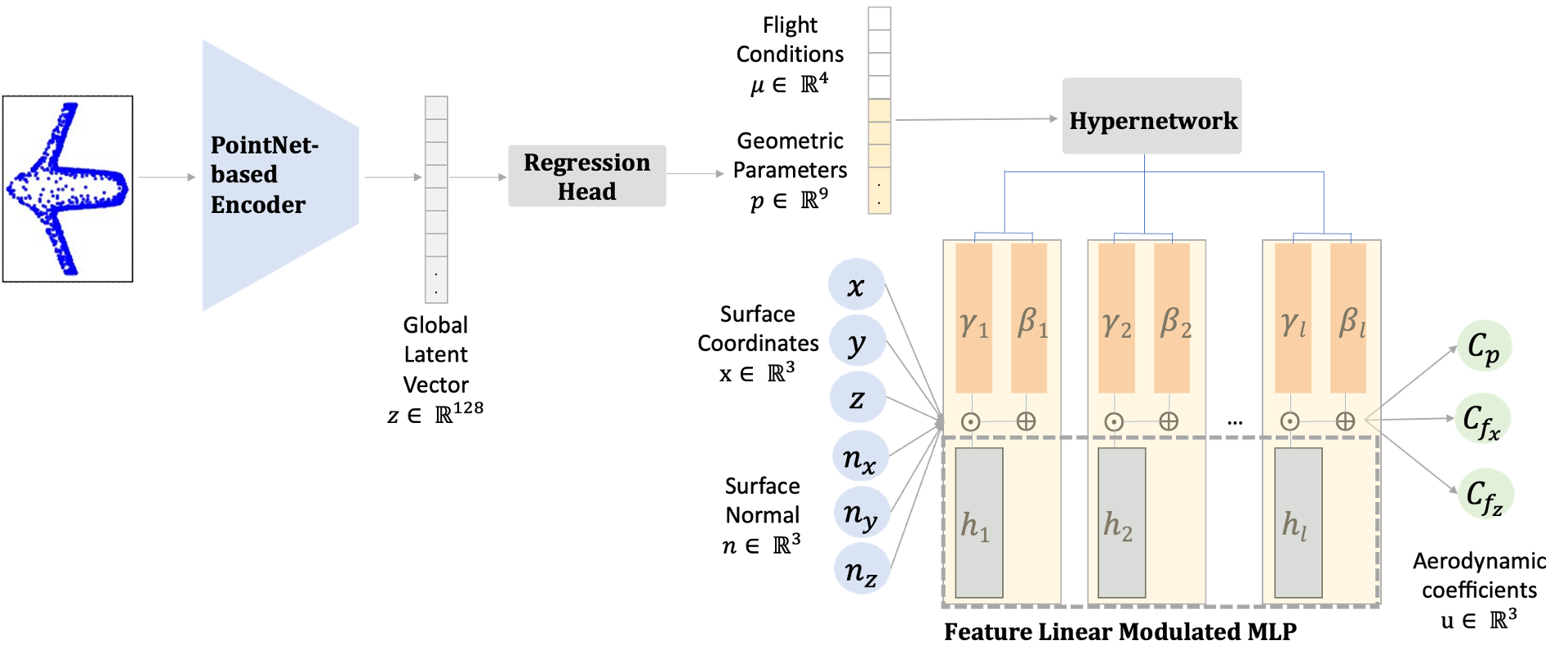}
    \caption{Overall architecture of the surrogate model. The network consists of two main components: (1) a PointNet-based encoder that maps sampled surface points to global geometric design parameters, and (2) a FiLM-based network that uses these parameters along with flight conditions to predict pointwise aerodynamic surface properties such as pressure and friction coefficients.}
    \label{fig:overall_model}
\end{figure}

The main goal of this model is to predict geometric design parameters directly from sampled surface points. \autoref{fig:overall_model} illustrates the architecture of the model.

We employ random sampling to subsample each point cloud into 15 batches, each containing 2048 representative points. Random sampling is computationally efficient and easily parallelizable, significantly reducing runtime and memory overhead when handling large aerodynamic point clouds.

To predict geometric shape parameters from these sampled points, we use a permutation-invariant neural network architecture inspired by PointNet~\cite{qi2017pointnet}. The architecture can be summarized mathematically as follows:

Given a point cloud \( X \in \mathbb{R}^{2048 \times 3} \), we first transform it through a shared multilayer perceptron (MLP):
\begin{equation}
X' = \phi(X), \quad X' \in \mathbb{R}^{2048 \times 128},
\end{equation}
where \(\phi\) denotes the shared MLP layers applied to each point independently.

Next, we apply a permutation-invariant global max-pooling operation to aggregate individual point features into a global latent vector \( z \):
\begin{equation}
z = \max(X'), \quad z \in \mathbb{R}^{128}.
\end{equation}

Finally, a regression head \( f_\theta \) maps the global latent vector \( z \) to the predicted geometric design parameters \(\hat{\mathbf{p}}\):
\begin{equation}
\hat{\mathbf{p}} = f_\theta(z), \quad \hat{\mathbf{p}} \in \mathbb{R}^{9}.
\end{equation}

The entire model is trained using a mean squared error (MSE) loss against ground-truth design parameters.

\subsection*{Feature-wise Linear Modulation (FiLM) Network for Predicting Pointwise Aerodynamic Surface Properties}

The main goal of the FiLM network is to predict pointwise aerodynamic surface properties conditioned on geometric shape parameters and flight conditions. Our use of FiLM is inspired by Catalani et al.~\cite{catalani2024neural}, who apply implicit neural representations (INRs) for aerodynamic field prediction using shift-only modulation based on flow conditions and 3D coordinates. While their approach relies on a two-stage meta-learning setup with shape-specific latent vectors learned through signed distance function (SDF) and pressure field reconstruction, our method directly conditions the FiLM network using \emph{explicit geometric design parameters} learnt from the PointNet model and known flight conditions, without relying on learned latent vectors. This leads to improved interpretability and eliminates the need for shape-specific latent optimization. Additionally, we use a more expressive FiLM architecture that performs \emph{both scaling and shifting} at each layer. Finally, our surrogate model predicts a richer set of aerodynamic surface quantities, including pressure coefficient (\(C_p\)) and skin friction coefficients (\(C_{f_x}, C_{f_z}\)), allowing for more detailed aerodynamic force modeling across the aircraft surface.

This network is trained using ground-truth geometric parameters, ensuring high accuracy and robustness in aerodynamic predictions. The surrogate model can operate in two distinct modes: if the geometric shape parameters are readily available, the user can directly input them into the FiLM network along with flight conditions; alternatively, if these parameters are not explicitly known, the permutation-invariant PointNet model can first predict the geometric parameters from sampled point clouds, and the predicted parameters are subsequently used to condition the FiLM network.

Our surrogate model predicts pointwise aerodynamic coefficients \( (C_p, C_{f_x}, C_{f_z}) \) by using a FiLM approach~\cite{perez2018film} conditioned on  geometric design parameters and flight conditions. FiLM is closely related to hypernetwork conditioning strategies~\cite{ha2016hypernetworks}, where a smaller network generates dynamic layer parameters (scale and shift) based on external inputs 
(e.g., geometric design parameters and flight conditions). This enables the surrogate to \emph{dynamically adapt} its internal representations for varying shape and flow conditions, thereby enhancing prediction accuracy.

Given input surface coordinates \(\mathbf{x}\in\mathbb{R}^3\), corresponding surface normals \(\mathbf{n}\in\mathbb{R}^3\), aerodynamic flight parameters \(\boldsymbol{\mu}\), and geometric design parameters \(\mathbf{p}\in\mathbb{R}^9\), the FiLM network \(f_{\theta}\) predicts the aerodynamic coefficients \(\mathbf{u}(\mathbf{x},\mathbf{n})\) defined as:

\[
\mathbf{u}(\mathbf{x},\mathbf{n})
\;=\;
\bigl[\,C_p(\mathbf{x},\mathbf{n}),\;C_{f_x}(\mathbf{x},\mathbf{n}),\;C_{f_z}(\mathbf{x},\mathbf{n})\,\bigr],
\]

via:

\[
f_{\theta}\bigl(\mathbf{x},\mathbf{n};\,\mathbf{p},\boldsymbol{\mu}\bigr)
\;=\;
\mathrm{Modulated\text{-}MLP}\bigl(\mathbf{x},\mathbf{n};\,\mathbf{p},\boldsymbol{\mu}\bigr)
\;\longrightarrow\;
\mathbf{u}(\mathbf{x},\mathbf{n}).
\]

\noindent where:
\begin{itemize}
  \item \(\mathbf{x} = (x,y,z)\) are the 3-D coordinates of a surface point.
  \item \(\mathbf{n} = (n_x,n_y,n_z)\) is the corresponding unit surface normal.
  \item \(\mathbf{p}\) are the nine geometric design parameters predicted by PointNet.
  \item \(\boldsymbol{\mu}\) are the flight conditions (Mach number, angle of attack, altitude, Reynolds length, etc.).
  \item \(C_p(\mathbf{x},\mathbf{n})\) is the pressure coefficient at the given point and normal.
  \item \(C_{f_x}(\mathbf{x},\mathbf{n}),\,C_{f_z}(\mathbf{x},\mathbf{n})\) are the skin friction coefficients in the \(x\) and \(z\) directions at that point.
\end{itemize}

\paragraph{FiLM Conditioning}

FiLM applies affine transformations to intermediate MLP activations based on external inputs 
\((\mathbf{p}, \boldsymbol{\mu})\). Mathematically, for layer \(l\), this operation is defined as:
\begin{equation}
\eta_{l}(\mathbf{h}_{l})
~=\;
\boldsymbol{\gamma}_{l} \,\odot\, \mathbf{h}_{l} + \boldsymbol{\beta}_{l},
\label{eq:film}
\end{equation}
where \(\mathbf{h}_l\) is the hidden activation at layer \(l\), and 
\(\boldsymbol{\gamma}_{l}\), \(\boldsymbol{\beta}_{l}\) are the \emph{scale} and \emph{shift} vectors 
that depend on \((\mathbf{p}, \boldsymbol{\mu})\)~\cite{perez2018film,ha2016hypernetworks}. 
The symbol \(\odot\) denotes elementwise multiplication. 
\emph{This design directly modulates the layer activations to dynamically tune the network's response for each aerodynamic scenario.}

\paragraph{Hypernetwork for Modulation Parameters}

To generate the FiLM parameters, we employ a hypernetwork \(h_{\psi}\), parameterized by 
\(\psi\), which takes the geometric shape parameters \(\mathbf{p}\) and flight conditions \(\boldsymbol{\mu}\) 
as inputs:
\begin{equation}
\bigl[\boldsymbol{\gamma}_{l},\; \boldsymbol{\beta}_{l}\bigr]
\;=\;
h_{\psi}\bigl(\mathbf{p},\, \boldsymbol{\mu}\bigr).
\end{equation}
Here, \(h_{\psi}\) is implemented as a smaller MLP that \emph{predicts} distinct scale and shift 
parameters \((\boldsymbol{\gamma}_l,\boldsymbol{\beta}_l)\) for each layer \(l\). 
\emph{Conditioning the surrogate network in this manner ensures that geometry and operating 
conditions can flexibly alter intermediate representations, thereby improving the capacity to handle 
complex, shape-dependent aerodynamic responses.}

\paragraph{Training Objective}

We jointly optimize the FiLM network and hypernetwork parameters \((\theta, \psi)\) by minimizing the mean squared error between the predicted aerodynamic coefficients \(\mathbf{u}_i(\mathbf{x},\mathbf{n})\) and the ground-truth CFD data across all sampled surface points and normals:
\[
\min_{\theta,\psi}
\;
\sum_{i=1}^{M}
\sum_{(\mathbf{x},\mathbf{n}) \in \Sigma_i}
\Bigl\|
   f_{\theta}\bigl(\mathbf{x},\mathbf{n}; \mathbf{p}_i,\boldsymbol{\mu}_i\bigr)
   - 
   \mathbf{u}_i(\mathbf{x},\mathbf{n})
\Bigr\|^{2}.
\]

Here, \(\Sigma_i\) denotes the set of coordinate–normal pairs for the \(i\)-th sample, and \(\mathbf{u}_i(\mathbf{x},\mathbf{n})\) is the ground-truth vector of aerodynamic coefficients at that point. In practice, the trained FiLM network can then predict the full surface field for new geometries and flight conditions by generating the appropriate modulation parameters.

\section{Model Training and Evaluation}
\paragraph{Training Procedure} 
Both PointNet and FiLM models were trained on aerodynamic configurations while maintaining a geometry-based data split. Specifically, 8,830 cases from 999 geometries, each sampled under different flight conditions, were grouped by geometry to ensure that cases from the same geometry remained within the same subset. These cases were divided into a 90\% training set and a 10\% validation set based on geometry. A separate test set consisting of 870 cases from 100 entirely distinct geometries was used to evaluate the model's performance. We used the Adam optimizer with a learning rate decay schedule to improve convergence and reduce overfitting.

\section{Results and Discussion}

\paragraph{PointNet Model Performance}
The PointNet model accurately predicts geometric shape parameters, demonstrating high correlation with ground truth parameters as quantified by the coefficient of determination ($R^2$), shown in ~\autoref{tab:pointnet_r2}.

\begin{table}[h]
    \centering
    \caption{PointNet Regressor $R^2$ Scores for Geometric Parameters}
    \label{tab:pointnet_r2}
    \begin{tabular}{@{} c c @{}}
        \hline
        \textbf{Design Parameter} & \textbf{$R^2$} \\
        \hline
        C2/C1 & 0.9893 \\
        C3/C1 & 0.9896 \\
        C4/C1 & 0.9945 \\
        B1/C1 & 0.9923 \\
        B2/C1 & 0.9948 \\
        B3/C1 & 0.9997 \\
        S1 & 0.9968 \\
        S2 & 0.9914 \\
        S3 & 0.9973 \\
        \hline
    \end{tabular}
\end{table}

\paragraph{FiLM Network Performance}
The FiLM network's detailed aerodynamic coefficient prediction performance is summarized in Table~\ref{tab:film_results}, comparing two distinct conditioning modes: (1) using ground truth geometric shape parameters and (2) using predicted shape parameters obtained from the PointNet model.

\begin{table}[h!]
    \centering
    \caption{FiLM Network Aerodynamic Coefficient Prediction Errors}
    \setlength{\tabcolsep}{12pt}
    \renewcommand{\arraystretch}{1.3}
    \begin{tabular}{lccc}
        \hline
        \textbf{Metric} & $C_p$ & $C_{f_x}$ & $C_{f_z}$ \\
        \hline
        \multicolumn{4}{c}{\textit{Conditioned on Ground Truth Parameters}} \\
        MSE & 7.86e-03 & 2.80e-05 & 1.51e-05 \\
        MAE & 3.72e-02 & 1.35e-03 & 7.96e-04 \\
        Rel L1 (\%) & 13.52 & 22.09 & 30.01 \\
        Rel L2 (\%) & 3.11 & 7.74 & 18.79 \\
        \hline
        \multicolumn{4}{c}{\textit{Conditioned on Predicted Parameters}} \\
        MSE & 1.19e-02 & 1.82e-04 & 5.72e-05 \\
        MAE & 4.33e-02 & 1.98e-03 & 1.19e-03 \\
        Rel L1 (\%) & 14.99 & 24.03 & 31.53 \\
        Rel L2 (\%) & 4.24 & 16.78 & 21.84 \\
        \hline
    \end{tabular}
    \label{tab:film_results}
\end{table}

\begin{figure}[htbp]
    \centering
    \includegraphics[width=\textwidth]{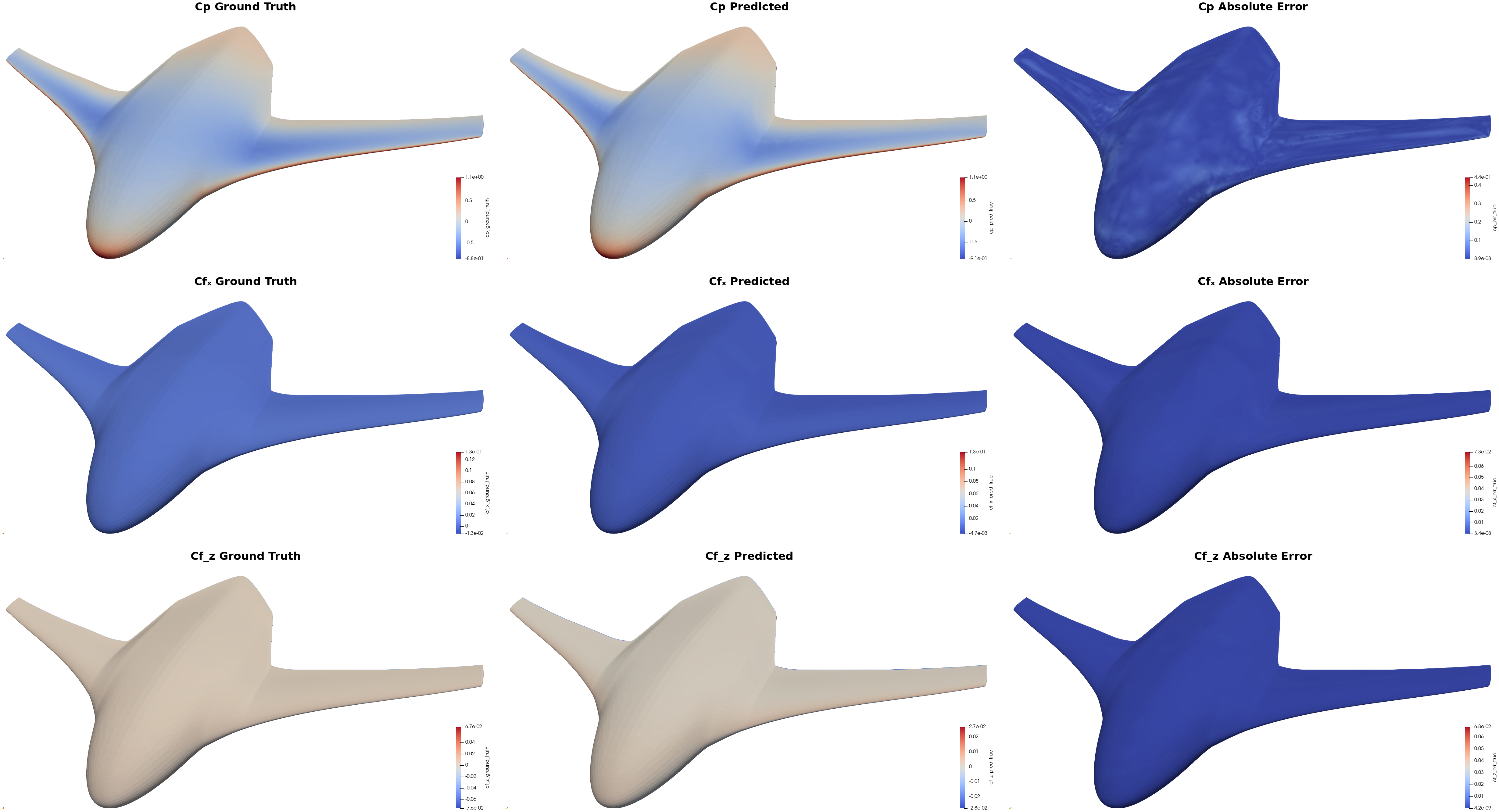}
    \caption{Comparison of \textbf{ground truth}, \textbf{predicted}, and \textbf{Absolute Error} on the aerodynamic coefficients for one test case generated by the FiLM model conditioned on predicted geometric design parameters. The rows correspond to the pressure coefficient ($C_p$) and the skin friction coefficients in the $x$ and $z$ directions ($C_{f_x}$ and $C_{f_z}$).}
    \label{fig:aerodynamic_prediction}
\end{figure}

We also integrate FiLM’s outputs over the surface to compute overall lift and drag. 
Even though the model is only trained on local pointwise fields, the summed results match closely with CFD ground truth, proving that this surrogate is indeed accurate.

\begin{figure}[htbp]
    \centering
    \includegraphics[width=0.9\textwidth]{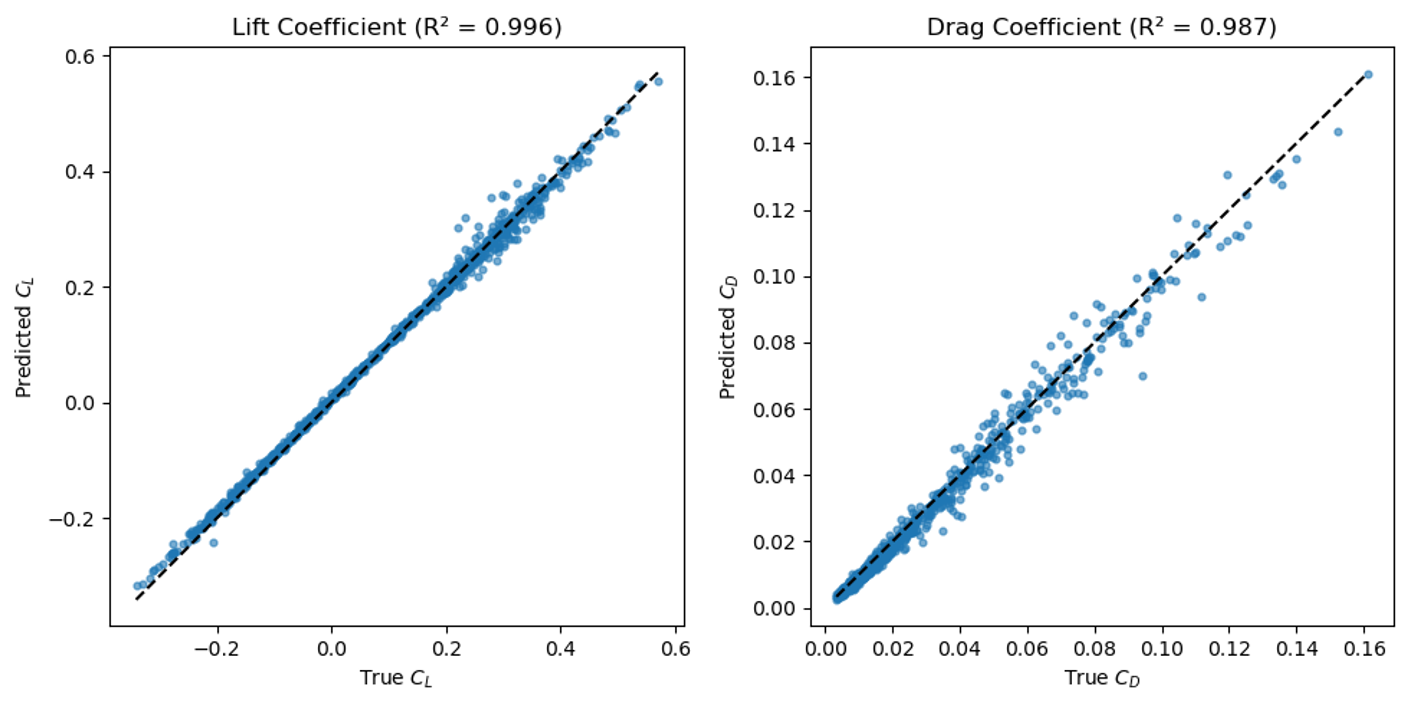}
    \caption{Correlation between predicted and true integrated aerodynamic coefficients. 
    The FiLM surrogate achieves $R^2 = 0.996$ for lift ($C_L$) and $R^2 = 0.987$ for drag ($C_D$), 
    confirming its accuracy in reproducing overall aerodynamic forces.}
    \label{fig:lift_drag_r2}
\end{figure}

\section{Conclusion, Limitations, and Future Work}

In this work, we introduced \textbf{BlendedNet}, a publicly available high-resolution dataset for Blended Wing Body (BWB) aerodynamic analysis, featuring 999 unique geometries with 8830 cases. This dataset was generated using Latin Hypercube Sampling (LHS) of geometric and flight parameters, followed by high-fidelity Reynolds-Averaged Navier-Stokes simulations with the Spalart-Allmaras turbulence model. 
To efficiently predict pointwise aerodynamic coefficients, we develop a two-stage deep learning surrogate modeling framework. First, we use a permutation-invariant PointNet model to infer geometric shape parameters directly from sampled aerodynamic surface point clouds. Subsequently, these predicted shape parameters, combined with known flight conditions, condition a Feature-wise Linear Modulation network to predict pointwise aerodynamic surface properties. Our results demonstrate that this framework accurately predicts surface aerodynamic properties with low error, presenting a promising data-driven alternative for accelerating BWB aerodynamic analysis. 

\begin{enumerate}
    \item \textbf{Infeasible or Unphysical Designs:} Some geometries generated through automated sampling may be aerodynamically unrealistic or structurally infeasible. 
    For instance, the lowest-drag design in our dataset is unlikely to be viable for practical aircraft development.
    This issue arises due to the nature of automated geometry generation, where the sampled design space is not explicitly constrained by manufacturability or stability considerations.
    \item \textbf{Surrogate Model Limitations:} While the FiLM-based surrogate model performs well in overall pointwise aerodynamic predictions, it struggles when there is a sharp difference in aerodynamic performance for neighboring points on the surface. Enhancing the expressiveness of the inputs by incorporating Gaussian and Fourier feature encodings could improve generalization and help better resolve these challenging aerodynamic features.
    \item \textbf{Absence of Full Flow-Field Predictions:} BlendedNet currently focuses on pointwise surface aerodynamic coefficients rather than full three-dimensional flow fields. Future work is planned to include flow-field solution data.
\end{enumerate}

\subsection*{Datasheet for BlendedNet Dataset}
\subsubsection*{Motivation}
\begin{itemize}
    \item \textbf{Why was the dataset created?} \\
    Provide a benchmark training dataset for the prediction of blended-wing body aircraft surface quantities.

    \item \textbf{Who created the dataset (team, individual)?} \\
    MIT Lincoln Laboratory and the MIT DeCoDe lab created the dataset.

    \item \textbf{Who funded the creation of the dataset?} \\
    This work was funded by MIT Lincoln Laboratory under Air Force Contract No. FA8702-15-D-0001.
\end{itemize}

\subsubsection*{Composition}
\begin{itemize}
    \item \textbf{What do the instances that comprise the dataset represent?} \\
    Surface quantities representing the pressure and skin friction forces on the aircraft for a particular geometry and flight condition. The dataset also includes integrated force coefficients (for each case) and the parameters used to define each geometry.

    \item \textbf{How many instances are there in total?} \\
    The training set contains 8{,}830 cases and the testing set contains 870  cases.

    \item \textbf{What data does each instance consist of?} \\
    Each sample contains:
    \begin{enumerate}

        \item Geometry parameters used to define the aircraft model
        \item Flight condition of the simulation (altitude, Mach number, Reynolds number, angle of attack, angle of side-slip)
        \item Integrated force coefficients ($C_L, C_D, C_{My}$)
        \item The aircraft model in VTK format, with distributed pressure and skin friction coefficients
    \end{enumerate}

    \item \textbf{Is there any missing data? If so, how much?} \\
    The dataset here is the complete set of CFD simulation cases which sufficiently converged.

    \item \textbf{How is the data associated with each instance organized?} \\
    The parameters for each geometry are stored in a configuration file, the flight condition and integrated force coefficients for each case are stored in a \texttt{.dat} file, and the surface quantities are stored in VTK format.
\end{itemize}

\subsubsection*{Collection Process}
\begin{itemize}
    \item \textbf{How was the data collected?} \\
    A data-generation pipeline of geometry creation, computational meshing, and computational fluid dynamics (CFD) simulation was used. Parametrically varied geometries were generated and simulated over many flight conditions using NASA FUN3D (RANS equations with the SA turbulence model).

    \item \textbf{Who was involved in the data collection process (e.g., annotators, researchers)?} \\
    Researchers at MIT Lincoln Laboratory.

    \item \textbf{Over what timeframe was the data collected?} \\
    Approximately three weeks in March 2025.

    \item \textbf{How was it decided which data to collect and which to exclude?} \\
    Convergence criteria were enforced for CFD simulations: density residuals of $10^{-5}$ and turbulence residuals of $10^{-4}$.
\end{itemize}

\subsubsection*{Preprocessing/cleaning/labeling}
\begin{itemize}
    \item \textbf{Was any preprocessing, cleaning, or labeling of the data done (e.g., normalization, annotation)? If so, how was this done and by whom?} \\
    No specific preprocessing beyond the standard CFD simulation outputs.
\end{itemize}

\subsubsection*{Uses}
\begin{itemize}
    \item \textbf{For what tasks in Engineering Design is the dataset suitable?} \\
    Development of models to predict integrated and distributed surface aerodynamic coefficients for blended-wing body aircraft, as well as design optimization of blended-wing body aircraft.

    \item \textbf{Has the dataset been used for any tasks already? If so, which ones?} \\
    The dataset has been used as described in the associated paper (e.g., for training and validating predictive models).

    \item \textbf{What performance metrics are relevant for assessing tasks using this dataset?} \\
    Prediction error on integrated force coefficients, distributed surface pressure, and skin friction (MSE, MAE, L1/L2 relative error).
\end{itemize}

\subsubsection*{Distribution}
\begin{itemize}
    \item \textbf{How is the dataset distributed?} \\
    The BlendedNet dataset is hosted on Harvard Dataverse and can be accessed at: 
\href{https://doi.org/10.7910/DVN/VJT9EP}{https://doi.org/10.7910/DVN/VJT9EP}.

    \item \textbf{Are there any restrictions or licenses on its use?} \\
    An MIT License is being considered.

\end{itemize}

\subsubsection*{Maintenance}
\begin{itemize}
    \item \textbf{Who is responsible for dataset maintenance?} \\
    MIT Lincoln Laboratory and the MIT DeCoDe lab 

    \item \textbf{How can individuals submit corrections or updates?} \\
    Contact the authors.

    \item \textbf{Is there a versioning system in place? If so, what is it?} \\
    Yes. Version control is implemented in Harvard dataverse.
\end{itemize}
DISTRIBUTION STATEMENT A. Approved for public release. Distribution is unlimited.
This material is based upon work supported under Air Force Contract No. FA8702-15-D-0001. Any opinions, findings, conclusions or recommendations expressed in this material are those of the author(s) and do not necessarily reflect the views of the U.S. Air Force.

\bigskip
\noindent\textcopyright~2025 Massachusetts Institute of Technology.
Delivered to the U.S. Government with Unlimited Rights, as defined in DFARS Part 252.227-7013 or 7014 (Feb 2014).

\bibliographystyle{unsrtnat}
\bibliography{template}

\clearpage
\appendix
\onecolumn
\section*{Appendix}

\begin{figure}[ht]
    \centering
    \includegraphics[width=\textwidth]{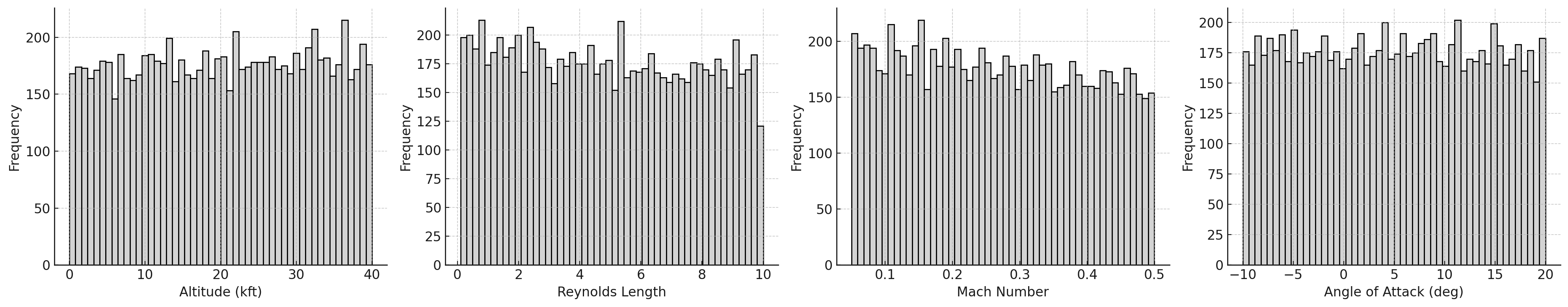}
    \caption{Distribution of flight condition parameters used in dataset generation. Latin Hypercube Sampling (LHS) was applied to sample altitude, Mach number, angle of attack, and Reynolds length.}
    \label{fig:flight_conditions}
\end{figure}

\begin{figure}[ht]
    \centering
    \includegraphics[width=0.8\textwidth]{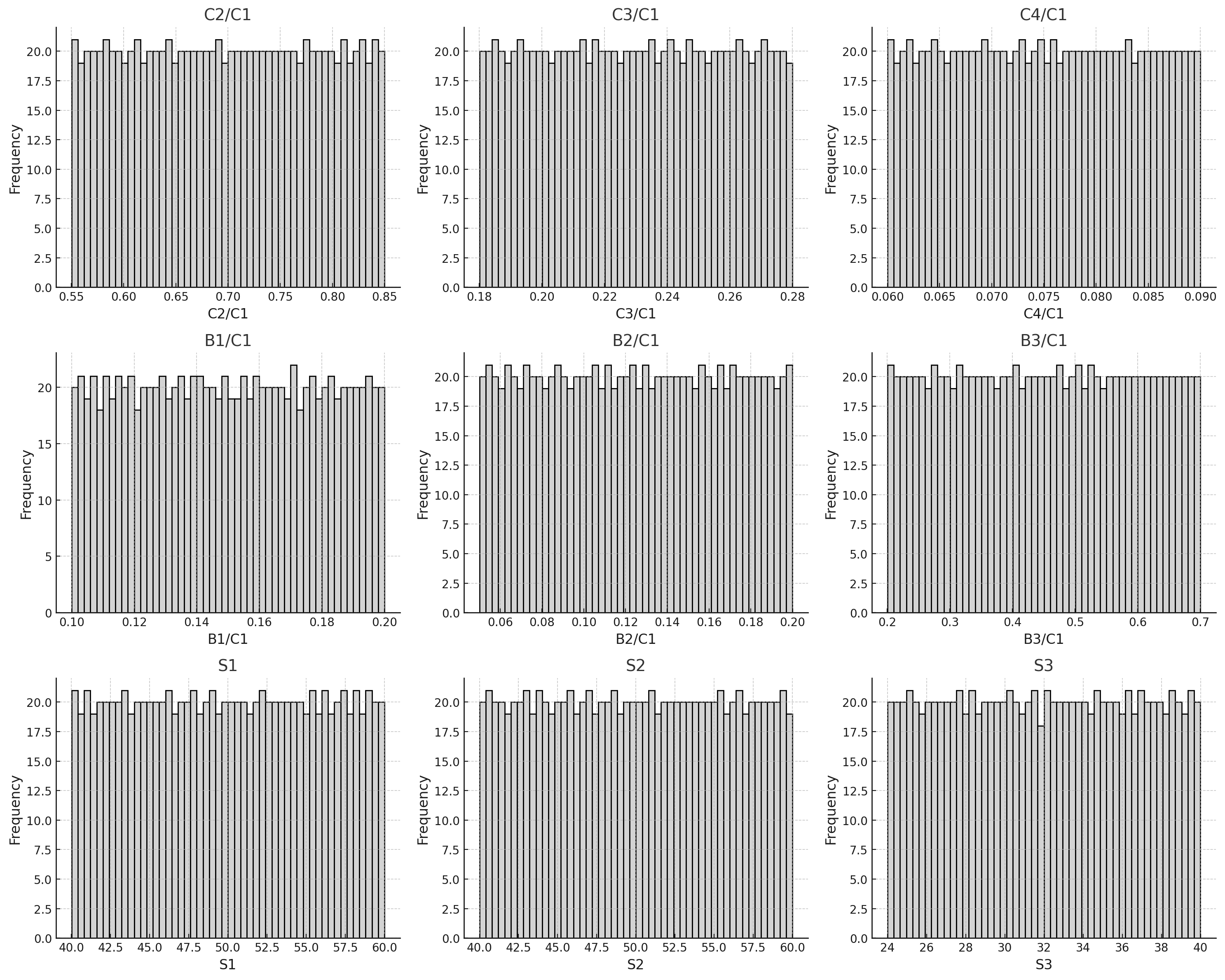}
    \caption{Distribution of geometric design parameters for the blended wing body (BWB) aircraft planform. Parameters include relative chord lengths (C2/C1, C3/C1, C4/C1), relative spanwise widths (B1/C1, B2/C1, B3/C1), and sweep angles (S1, S2, S3). These parameters were sampled using Latin Hypercube Sampling (LHS) to ensure a well-spread design space.}
    \label{fig:geom_param_dist}
\end{figure}

\twocolumn

\end{document}